# A Discovery Algorithm for Directed Cyclic Graphs

Thomas Richardson
Philosophy Department
Carnegie-Mellon University
Pittsburgh, PA 15213
e-mail: tsr+@andrew.cmu.edu

## Abstract

Directed acyclic graphs have been used fruitfully to represent causal structures (Pearl 1988). However, in the social sciences and elsewhere models are often used which correspond both causally and statistically to directed graphs with directed cycles (Spirtes 1995). Pearl (1993) discussed predicting the effects of intervention in models of this kind, so-called linear non-recursive structural equation models. This raises the question of whether it is possible to make inferences about causal structure with cycles, from sample data. In particular do there exist general, informative, feasible and reliable procedures for inferring causal structure from conditional independence relations among variables in a sample generated by an unknown causal structure? In this paper I present a discovery algorithm that is correct in the large sample limit, given commonly (but often implicitly) made plausible assumptions, and which provides information about the existence or non-existence of causal pathways from one variable to another. The algorithm is polynomial on sparse graphs.

## 1 DIRECTED GRAPH MODELS

A Directed Graph $G$ consists of an ordered pair $<V,E>$ where $V$ is a set of vertices, and $E$ is a set of directed edges between vertices.[1] If there are no directed cycles[2] in $E$, then $<V,E>$ is called a Directed Acyclic Graph or (DAG). A Directed Cyclic Graph (DCG) *model* (Spirtes 1995) is an ordered pair $<G,\mathcal{P}>$ consisting of a directed graph $G$ (cyclic or acyclic) and a joint probability distribution $\mathcal{P}$ over the set $V$ in which certain conditional independence relations, encoded by the graph, are true.[3] Directed Acyclic Graph (DAG) models correspond to the special case in which $G$ is acyclic. The independencies encoded by a given graph are determined by a graphical criterion called d-separation, as explained for the acyclic case in Pearl (1988), and extended to the cyclic case in Spirtes (1995) (See also Koster 1994). The following definition can be applied to cyclic and acyclic cases and is equivalent to Pearl's in the latter:

**Definition:** d-connection/d-separation for directed graphs
For disjoint sets of vertices, $X$, $Y$ and $Z$, $X$ is *d-connected to* $Y$ *given* $Z$ if and only if for some $X \in X$, and $Y \in Y$,[4] there is an (acyclic) undirected path $U$ between $X$ and $Y$, such that:

(i) If there is an edge between A and B on U, and an edge between B and C on U, and $B \in Z$, then B is a collider between A and C relative to U, i.e. $A \rightarrow B \leftarrow C$ is a subpath of U.

(ii) If B is a collider between A and C relative to U, then there is a descendant D,[5] of C, and $D \in Z$.

For disjoint sets of vertices, $X$, $Y$ and $Z$, if $X$ and $Y$ are not d-connected given $Z$, then $X$ and $Y$ are said to be *d-separated* given $Z$.

The constraint relating $G$ and $\mathcal{P}$ in a DCG model $<G,\mathcal{P}>$ is:

### 1.1 The Global Directed Markov Condition

A DCG model $<G,\mathcal{P}>$ is said to satisfy the Global Directed Markov Property if for all disjoint sets of variables $A$, $B$ and $C$, if $A$ is d-separated from $B$ given $C$ in $G$, then $A \perp\!\!\!\perp B \mid C$ in $\mathcal{P}$.[6]

This condition is important since a wide range of statistical models can be represented as DAG models satisfying the Global Directed Markov Condition, including recursive linear structural equation models with independent errors, regression models, factor analytic models, and discrete latent variable models (via extensions of the formalism). An alternative, but equivalent, definition is given by Lauritzen *et al.* (1990).

---

[1] If $<A,B> \in E$, A, B distinct, then there is said to be an edge *from* A *to* B, represented by $A \rightarrow B$. If $<A,B> \in E$ *or* $<B,A> \in E$, then in either case there is said to be an edge *between* A and B.
[2] By a 'directed cycle' I mean a directed path $X_0 \rightarrow X_1 \ldots \rightarrow X_{n-1} \rightarrow X_0$ of n distinct vertices, where $n \geq 2$.
[3] Since the elements of $V$ are both vertices in a graph, and random variables in a joint probability distribution the terms 'variable' and 'vertex' can be used interchangeably.
[4] Upper case Roman letters (V) are used to denote sets of variables, and plain face Roman letters (V) to denote single variables. |V| denotes the cardinality of the set V.
[5] 'Descendant' is defined as the reflexive, transitive closure of the 'child' relation, hence every vertex is its own descendant. Similarly every vertex is its own ancestor.
[6] '$X \perp\!\!\!\perp Y \mid Z$' means that 'X is independent of Y given Z'.

However, not all models can be represented thus as DAG models. Spirtes (1995) has shown that the conditional independencies which hold in non-recursive linear structural equation models[7] are precisely those entailed by the Global Directed Markov condition, applied to the cyclic graph naturally associated with a non-recursive structural equation model[8] with independent errors. It can be shown that in general there is no DAG encoding the conditional independencies which hold in such a model. Non-recursive structural equation models are used to model systems with feedback, and are applied in sociology, economics, biology, and psychology.

We make two assumptions connecting the probability distribution $P$ and the true causal graph $G$:

### The Causal Markov Assumption

A distribution generated by a causal structure represented by a directed graph $G$ satisfies the Global Directed Markov condition.

For linear structural equation models this is true by definition if the error terms are independent.

### The Causal Faithfulness Assumption

All conditional independence relations present in $P$ are consequences of the Global Directed Markov condition applied to the true causal structure $G$.

This is an assumption that any conditional independence relation true in $P$ is true in virtue of causal structure rather than a particular parameterization of the model. (For further justification and discussion see Spirtes et al. 1993)

## 2 DISCOVERY

(Cyclic or Acyclic) graphs $G_1$ and $G_2$ are *Markov equivalent* if any distribution which satisfies the Global Directed Markov condition with respect to one graph satisfies it with respect to the other, and vice versa. The class of graphs which are Markov Equivalent to $G$ is denoted Equiv($G$).

It can be shown to follow from the fact that the Global Directed Markov condition only places conditional independence constraints on distributions, that, under this definition, two graphs are Markov equivalent if and only if the same d-separation relations hold in both graphs.

### 2.1 THE DISCOVERY PROBLEM

Given an oracle for conditional independencies in a distribution $P$, satisfying the Global Markov and Faithfulness conditions w.r.t. some directed (cyclic or acyclic) graph $G$ without hidden variables, is there an efficient, reliable algorithm for making inferences about the structure of $G$?

Since if $P$ satisfies the Global Markov and Faithfulness conditions w.r.t. to $G$, then it also satisfies them w.r.t. every graph $G^*$ in Equiv($G$), conditional independencies cannot be used to distinguish between graphs in Equiv($G$). Thus a procedure solving the Discovery Problem will determine causal features common to all graphs in a given Markov equivalence class Equiv($G$), given an oracle for conditional independencies in $P$.

I present a feasible (on sparse graphs) algorithm which outputs a list of features common to all graphs in Equiv($G$), given an oracle for conditional independence relations in a distribution $P$, satisfying the Global Markov and Faithfulness conditions w.r.t. some directed (cyclic or acyclic) graph $G$. The strategy adopted is to construct a graphical object, called a Partial Ancestral Graph (PAG) which represents features common to all graphs in the Markov equivalence class (See Figure 1).

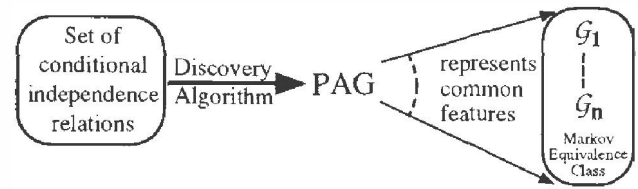

Figure 1: Strategy For Discovery Algorithm

### 2.2 PARTIAL ANCESTRAL GRAPHS (PAGs)

A PAG consists of a set of vertices **V**, a set of edges between vertices, and a set of edge-endpoints, two for each edge, drawn from the set $\{o, -, >\}$. In addition, pairs of edge endpoints may be connected by underlining, or dotted underlining. In the following definition '*' is a meta-symbol indicating the presence of any one of $\{o, -, >\}$.

**Definition:** Partial Ancestral Graph (PAG)

$\Psi$ is a PAG for Directed Cyclic Graph $G$ with vertex set **V**, if and only if

(i) There is an edge between A and B in $\Psi$ if and only if A and B are d-connected in $G$ given all subsets $\mathbf{W} \subseteq \mathbf{V}\setminus\{A,B\}$.

(ii) If there is an edge in $\Psi$, A—∗B, marked with a '–' at A, then in every graph in Equiv($G$), A is an ancestor of B.

(iii) If there is an edge in $\Psi$, A∗–>B, marked with a '>' at B, then in every graph in Equiv($G$), B is **not** an ancestor of A.

(iv) If there is an underlining A∗—∗B∗—∗C in $\Psi$, then B is an ancestor of (at least one of) A or C in every graph in Equiv($G$).

(v) If there is an edge from A to B, and from C to B, (A—>B<—C), then the arrow heads at B in $\Psi$ are joined by dotted underlining, thus A—>B<—C, only if in every graph in Equiv($G$) B is not a descendant of a common child of A and C.

(vi) Any edge endpoint not marked in one of the above ways is left with a small circle thus: o—∗.

---

[7] A non-recursive structural equation model is one in which the matrix of coefficients not fixed at zero is not lower triangular, for any ordering of the equations. (Bollen 1989)

[8] i.e. the directed graph in which X is a parent of Y, if and only if the coefficient of X in the structural equation for Y is not fixed at zero by the model.

Condition (i) differs from the other five conditions in stating necessary *and* sufficient conditions for a symbol, an edge, to appear in a PAG; (ii)-(vi) merely give necessary conditions. For this reason there are in fact many different PAGs for a graph $G$, though they all have the same edges, though not necessarily endpoints. Some of the PAGs provide more information than others about causal structure, e.g. they have fewer 'o's at the end of edges.[9] Some PAGs (providing less information) represent graphs from different Markov equivalence classes. However, the PAGs output by the discovery algorithm I present provide sufficient information so as to ensure that graphs with the features described by a particular PAG all lie in one Markov equivalence class. By the definition of a PAG, if $\Psi$ is a PAG for $G$, then $\Psi$ is also a PAG for every $G^* \in \text{Equiv}(G)$. Hence, a PAG $\Psi$ produced by the algorithm represents a unique Markov equivalence class.

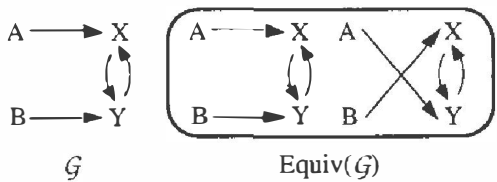

Figure 2: Graph $G$ and Equiv($G$)

Consider the graph $G$ in Figure 2. This graph entails that $A \perp\!\!\!\perp B$, and $A \perp\!\!\!\perp B \mid \{X,Y\}$ in any distribution $P$ with respect to which it satisfies the Global Directed Markov condition. In this case it can be proved that Equiv($G$) includes only the two graphs shown.[10] Figure 3 shows the PAG given by the algorithm I give, given a conditional independence oracle for a distribution $P$ satisfying the Global Directed Markov and Faithfulness w.r.t. $G$.

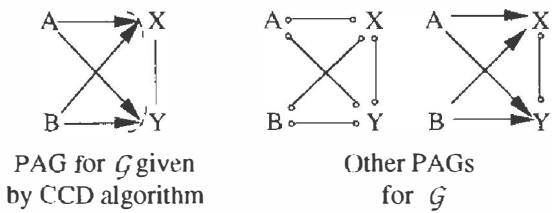

PAG for $G$ given         Other PAGs
by CCD algorithm          for $G$

Figure 3: PAGs for Graph $G$ in Fig.2

The PAG given by the algorithm allows us to make the following inferences (among others) about every graph in Equiv($G$), and hence about $G$:

(a) X is an ancestor of Y, and vice versa, so there is a cycle between X and Y.

(b) Neither X nor Y is an ancestor of A or B.

(c) Both A and B are ancestors of X and Y.

Note that not every edge in the PAG appears in every graph in Equiv($G$), e.g. in the example in Figure 2. This is because an edge in the PAG indicates only that the two variables connected by the edge are d-connected given any subset of the other variables. In fact it is possible to show that if there is an edge between two vertices in a PAG, then there is a graph represented by the PAG in which that edge is present. The algorithm I present does not always give the most informative PAG for a given graph $G$; there may be features common to all graphs in the Markov equivalence class which are not captured by the PAG the algorithm outputs. In this sense the algorithm is not complete, though the algorithm is 'd-separation complete' in the sense that each PAG it outputs represents a unique Markov equivalence class.

## 3 CYCLIC CAUSAL DISCOVERY (CCD) ALGORITHM

Two vertices, X, Y in a PAG are *adjacent* if there is an edge between them, i.e. $A*\!-\!*B$. **Adjacent**($\Psi$,X) is the set of vertices adjacent to X in PAG $\Psi$.[11]

**Input:** A conditional independence oracle for a distribution $P$, satisfying the Global Directed Markov and Faithfulness conditions w.r.t. a directed (cyclic or acyclic) graph $G$ with vertex set **V**.

(In practice of course statistical tests of conditional independence in sample data take the place of the conditional independence oracle.)

**Output:** A PAG $\Psi$ for the Markov equivalence class Equiv($G$).

¶A Form a PAG $\Psi$ with an edge Xo—oY between every pair of vertices in **V**.
$n = 0$
repeat
   repeat
      Select an ordered pair of variables X and Y that are adjacent in $\Psi$ s.t. |**Adjacent**($\Psi$,X)\\{Y}| $\geq n$, and a set $S \subseteq$ **Adjacent**($\Psi$,X)\\{Y} s.t. |S| $=n$..
      If $X \perp\!\!\!\perp Y \mid S$, delete edge Xo-oY from $\Psi$ and record **S** in **Sepset**(X,Y) and **Sepset**(Y,X).[12]
   until all pairs of adjacent variables X,Y such that |**Adjacent**($\Psi$,X)\\{Y}| $\geq n$ and all sets $S \subseteq$ |**Adjacent**($\Psi$,X)\\{Y}| s.t. |S|$=n$ have been tested.
   $n = n+1$;
until for all ordered pairs of adjacent vertices X,Y, |**Adjacent**($\Psi$,X)\\{Y}|<$n$

¶B. For each triple of vertices A,B,C s.t. the pair A,B and the pair B,C are each adjacent in $\Psi$ but the pair A,C are not adjacent in $\Psi$, orient $A*\!-\!*B*\!-\!*C$ as A—>B<—C if and only if B $\notin$ **Sepset**<A,C>; orient $A*\!-\!*B*\!-\!*C$ as $A*\!-\!\underline{*B*}\!-\!*C$ if and only if B $\in$ **Sepset**<A,C>.

---

[9] If one PAG has a '>' at the end of an edge, then every other PAG for the same graph either has a '>' or a 'o' in that location. Similarly if one PAG has a '—' at the end of an edge then every other PAG either has a '—' or an 'o' in that location.

[10] Note that adding the A→Y, and/or B→X edges to $G$ produces a graph that is *not* Markov Equivalent to $G$; with these edges added A and B are no longer d-separated given {X, Y}, though they are in $G$.

[11] Again '*' is a meta-symbol indicating any of the three ends -, o, >.

[12] **Adjacent**($\Psi$,X) is updated when the graph $\Psi$ changes during ¶A. So Y$\notin$ **Adjacent**($\Psi$,X), X$\notin$ **Adjacent**($\Psi$,Y), after the edge Xo—oY is removed.

¶

¶C. For each triple of vertices <A,X,Y> in Ψ such that (a) A is not adjacent to X or Y, (b) X and Y are adjacent, (c) X ∉ **Sepset**<A,Y> then orient X *−*Y as X<—Y if A ⊥ X | **Sepset**<A,Y>.

¶D. For each vertex V in Ψ form the following set: X∈ **Local**(Ψ,V) ⇔ X is adjacent to V in Ψ, or there is a vertex Y s.t. X—>Y<—V in Ψ.[13]

$m = 0$

  repeat

    repeat

select an ordered triple <A,B,C> such that A—>B<—C, A and C are not adjacent, and |**Local**(Ψ,A)\{B,C}| ≥ $m$, and a set T ⊆ **Local**(Ψ,A)\{B,C}, |T|=$m$, and if A⊥C | T∪{B} then orient A—>B<—C as A—>B≤—C, and record T ∪ {B} in **Supset**<A,B,C>.

until for all triples such that A—>B<—C, (not A—>B≤—C), A and C are not adjacent, |**Local**(Ψ,A)\{B}| ≥ $m$, every subset T ⊆ **Local**(Ψ,A), |T| = $m$ has been considered.

$m = m + 1$;

until for all ordered triples <A,B,C> s.t. A—>B<—C, A and C not adjacent, are such that |**Local**(Ψ,A)\{B}| < $m$.

¶E. If there is a quadruple <A,B,C,D> of distinct vertices in Ψ such that (i) A—>B≤—C, (ii) A—>D<—C or A—>D≤—C, (iii) B and D are adjacent, then orient B*−*D as B—>D in Ψ if D ∉ **Supset**<A,B,C>

else orient B*−*D as B*—D.

¶F. For each quadruple <A,B,C,D> in Ψ of distinct vertices s.t. D is not adjacent to both A and C, and A—>B≤—C, if A ⊥ D | **Supset**<A,B,C> ∪ {D}, then orient B*—*D as B—>D in Ψ

### 3.1 SOUNDNESS AND COMPLETENESS
**Theorem 1 (Soundness)**

Given as input a conditional independence oracle for a distribution $\mathcal{P}$, satisfying the Global Directed Markov and Faithfulness assumptions w.r.t. a (cyclic or acyclic) graph $\mathcal{G}$, the CCD algorithm outputs a PAG Ψ for $\mathcal{G}$.

The proof of Theorem 1 is given in §4.

**Theorem 2 (d-separation Completeness)**

If the CCD algorithm, when given as input conditional independence oracles for distributions $\mathcal{P}_1$, $\mathcal{P}_2$ satisfying the Global Directed Markov and Faithfulness w.r.t. graphs $\mathcal{G}_1$, $\mathcal{G}_2$ respectively, produces as output PAGs Ψ$_1$, Ψ$_2$ respectively, then Ψ$_1$ ≡ Ψ$_2$ if and only if $\mathcal{G}_1$ and $\mathcal{G}_2$ are Markov equivalent.

The proof, (Richardson, 1996) exploits the characterization of Markov equivalence in Richardson (1994) to establish that if $\mathcal{G}_1$ and $\mathcal{G}_2$ are not Markov equivalent then the algorithm produces different PAGs. (It follows directly from Theorem 1 that if $\mathcal{G}_1$ and $\mathcal{G}_2$ are equivalent then Ψ$_1$ ≡ Ψ$_2$.)

### 3.2 TRACE OF CCD ALGORITHM

If given a conditional independence oracle for $\mathcal{G}$ in figure 2 the algorithm runs as follows: (Steps ¶C and ¶F do not perform any orientations here.)

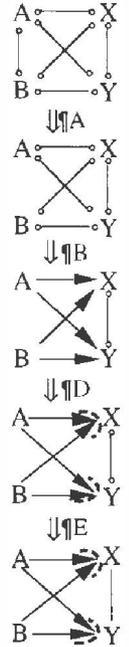

¶A: A ⊥ B | ∅ ⇒ A o–o B edge removed
    **Sepset**<A,B> = **Sepset**<B,A> = ∅

¶B: X ∉ **Sepset**<A,B> ⇒ A–>X<–B
    Y ∉ **Sepset**<A,B> ⇒ A–>Y<–B

¶D: A ⊥ B | {X,Y}
    **Supset**<A,X,B> = {X,Y} ⇒ A–>X≤–B
    **Supset**<A,Y,B> = {X,Y} ⇒ A–>Y≤–B

¶E: A–>X≤–B, A–>Y≤–B, X o–o Y,
    Y ∈ **Supset**<A,X,B> ⇒ X o—Y
    X ∈ **Supset**<A,Y,B> ⇒ X—Y

### 3.3 COMPLEXITY OF CCD ALGORITHM

Let $r = \underset{Y \in V}{\text{Max}} |\{X | Y \leftarrow X, \text{ or } X \leftarrow Y \text{ in } \mathcal{G}\}|$,

$k = \underset{Y \in V}{\text{Max}} |\{X | X \text{ is adjacent to Y in any PAG for } \mathcal{G}\}|$,[14]

and $n$ = no. of vertices in $\mathcal{G}$. It then follows that in searching (possibly unsuccessfully) for **Sepset**<X,Y> for every pair of distinct variables X,Y,

$$\text{Total no. of tests of conditional independence in ¶A} \leq 2 \cdot \binom{n}{2} \sum_{i=0}^{k} \binom{n-2}{i}$$

$$\leq \frac{(k+1)n^2(n-2)^{k+1}}{k!}.$$

Since $k \leq r^2$, this step is $O(n^{r^2+3})$. (Even as a worst case complexity bound this is loose.) ¶C performs at most one conditional independence test for each triple satisfying the conditions given, so this step is $O(n^3)$. In searching (possibly unsuccessfully) for sets **Supset**<X,Y,Z> for triples of distinct variables <X,Y,Z>

$$\text{Total no. of tests of conditional independence in ¶D} \leq 3 \cdot \binom{n}{3} \sum_{i=0}^{m} \binom{n-3}{i}$$

$$\leq \frac{(m+1)n^3(n-3)^{m+1}}{m!}$$

where $m = \underset{X \in V}{\text{Max}} |\textbf{Local}(\Psi,X)|$ in ¶D. Since $m \leq r^2$, it follows that ¶D is $O(n^{r^2+4})$. ¶F performs at most one test for each quadruple satisfying the conditions, so this

---

[13]Local(Ψ,A) is not recalculated as the algorithm progresses.

[14]Note k≠r since there may be an edge between two variables X*−*Y in a PAG for $\mathcal{G}$, even if there is no edge between X and Y in $\mathcal{G}$

step is $O(n^4)$. (¶B and ¶E do not perform any tests). Hence, the complexity of the algorithm is polynomial in the number of vertices for graphs of fixed degree (r); it is of course exponential in r. Although there are exponentially many conditional independence facts to check, the algorithm exploits entailment relations to obviate checking most of them when the graph is sparse.

# 4 Proof of Theorem 1 (Soundness)

The proof proceeds by showing that each section of the algorithm makes correct inferences from conditional independencies in $\mathcal{P}$, to the structure of any graph satisfying the Global Directed Markov and Faithfulness conditions w.r.t. to $\mathcal{P}$. If $\mathcal{P}$ satisfies these two conditions w.r.t. a graph $\mathcal{G}$, then $X \perp\!\!\!\perp Y \mid Z$, if and only if $X$ is d-separated from $Y$ by $Z$ in $\mathcal{G}$. Hence, the oracle for conditional independencies can be thought of as an oracle for testing d-separation relations in $\mathcal{G}$.

## 4.1 SECTION ¶A

**Lemma 1**: Given a PAG $\Psi$ for graph $\mathcal{G}$, if in $\mathcal{G}$ either (i) $X \rightarrow Y$ or (ii) $Y \leftarrow X$ or (iii) there is some vertex Z, s.t. $X \rightarrow Z \leftarrow Y$, and Z is an ancestor of X or Y (or both) *then* X and Y are adjacent in $\Psi$, i.e. X and Y are d-connected given any subset $S \subseteq V \setminus \{X,Y\}$ of the other vertices in $\mathcal{G}$.

**Proof**: If (i) holds then the path $X \rightarrow Y$ d-connects X and Y given any subset $S \subseteq V \setminus \{X,Y\}$, hence X and Y are adjacent in any PAG $\Psi$ for graph $\mathcal{G}$. The case in which (ii) holds is equally trivial: $X \leftarrow Y$ is a d-connecting path given any set $S \subseteq V \setminus \{X,Y\}$. If (iii) holds there is a common child (Z) of X and Y which is an ancestor of X or Y; therefore either there is a directed path $X \rightarrow Z \rightarrow A_1 \rightarrow \ldots A_n \rightarrow Y$ ($n \geq 0$), or there is a directed path $Y \rightarrow Z \rightarrow A_1 \rightarrow \ldots A_n \rightarrow X$. Suppose without much loss of generality that it is the former. Let S be an arbitrary subset of the other variables ($S \subseteq V \setminus \{X,Y\}$). If $S \cap \{Z, A_1 \ldots A_n\} \neq \emptyset$ then $X \rightarrow Z \leftarrow Y$ is a d-connecting path given S. If $S \cap \{Z, A_1 \ldots A_n\} = \emptyset$ then $X \rightarrow Z \rightarrow A_1 \rightarrow \ldots A_n \rightarrow Y$ is d-connecting given S. ∴

**Lemma 2**: In a graph $\mathcal{G}$, with vertices V, if all of the following hold:[15]

(i) X is not a parent of Y in $\mathcal{G}$

(ii) Y is not a parent of X in $\mathcal{G}$

(iii) there is no vertex Z s.t. Z is a common child of X and Y, and Z is an ancestor of X or Y

then for any set Q, X and Y are d-separated given T defined as follows:

$S = \text{Children}(X) \cap \text{Ancestors}(\{X,Y\} \cup Q)$

$T = [\text{Parents}(S \cup \{X\}) \cup S] \setminus$
$\quad [\text{Descendants (Children}(X) \cap \text{Children}(Y)) \cup \{X,Y\}]$

**Proof**: Every vertex in S is an ancestor of X or Y or Q. Every vertex in T is either a parent of X, a vertex in S, or a parent of a vertex in S, hence every vertex in T is an ancestor of X or Y or Q.

---

[15] i.e. None of the conditions in the antecedent of Lemma 1 hold.

**Claim**: If (i),(ii) and (iii) hold then X and Y are d-separated by T.

Suppose there is an undirected path P d-connecting X and Y. Let W be the first vertex on P. ((i) and (ii) imply $W \neq Y$.) There are two cases:

**Case 1** The path P goes $X \leftarrow W \ldots Y$.

**Subcase A**: W is not a descendant of a common child of X and Y. In this case $W \in T$ (Since W is a parent of X). Thus since W is a non-collider on P, P is not d-connecting given T. Contradiction.

**Subcase B**: W is a descendant of a common child of X and Y. In this case since X is a child of W, then X is a descendant of some common child Z of X and Y. But then Z is an ancestor of X, contradicting (iii).

**Case 2** The path P goes $X \rightarrow W \ldots Y$.

**Subcase A**: W is not a descendant of a common child of X and Y.

Let V be the next vertex on the path.

**Sub-subcase a**: The path P goes $X \rightarrow W \leftarrow V \ldots Y$.

If P is d-connecting, then some descendant of W is in T, but then some descendant of W is an ancestor of X or Y or Q. So W is an ancestor of X, Y or Q, hence $W \in S$. Moreover, since W is (by hypothesis) not a descendant of a common child, $V \neq Y$, and V is not a descendant of a common child of X and Y. V is a parent of W, $W \in S$, $X \neq V \neq Y$, so $V \in T$. Hence P fails to d-connect given T.

**Sub-subcase b**: The path P goes $X \rightarrow W \rightarrow V \ldots Y$.

If P d-connects given T, then W is either an ancestor of Y or some vertex in T. However, if W is an ancestor of some vertex in T, then W is an ancestor of X, Y or Q, so $W \in S$. Since W is (by hypothesis) not a descendant of a common child of X and Y, and $X \neq W \neq Y$, $W \in T$. Since in this case W occurs as a non-collider on P, P fails to d-connect given T. (This allows for the possibility that $V=Y$).

**Subcase B**: W is a descendant of a common child.

Thus Descendants (W) $\cap T = \emptyset$, since descendants of W are also descendants of common children of X and Y and so cannot occur in T.

Since no descendant of W is in T, if W occurs on d-connecting path P, then W is a non-collider on P. Suppose that there is a collider on P, take the first collider on the path after W, let us say <A,B,C>, so that P now takes the form: $X \rightarrow W \rightarrow \ldots \rightarrow \ldots \rightarrow A \rightarrow B \leftarrow C \ldots Y$. Since <A,B,C> is the first collider after W, it follows that B is a descendant of W. But if P is d-connecting then there is some descendant D of B, s.t. $D \in T$. But then since D is a descendant of B, and B is a descendant of W, $D \in$ Descendants (W) which is a contradiction since Descendants (W) $\cap T = \emptyset$. Hence every vertex on P is a non-collider.

As there are no colliders on P it follows that W is an ancestor of Y. But then W is a descendant of a common child of X and Y, *and* an ancestor of Y. But this contradicts (iii).

This completes the proof of Lemma 2. ∴

**Corollary A:** Given a graph $G$, and PAG $\Psi$ for $G$, X and Y are adjacent in $\Psi$ if and only if one of the following holds in $G$: (a) X is a parent of Y, (b) Y is a parent of X (c) There is some vertex Z which is a child of both X and Y, such that Z is an ancestor of either X or Y (or both)

**Proof:** 'If' is proved by Lemma 1, 'Only if' by Lemma 2 with $Q=\emptyset$ ∴

X and Y are said to be adjacent *in* $G$ if at least one of (a), (b), (c) holds for X,Y in $G$. By Corollary A, X and Y are adjacent in $G$ if and only if X and Y are adjacent in every PAG for $G$. Therefore I refer to a pair of variables as adjacent without specifying whether in a graph $G$ or a PAG for $G$.

**Corollary B:** In a graph $G$, if X and Y are d-separated by some set **R**, then X and Y are d-separated by a set **T** in which every vertex is an ancestor of X or Y. Further, either **T** is a subset of the vertices adjacent to X or X is an ancestor of Y.

**Proof:** Let **S**, **T** be the sets defined in Lemma 2 with $Q=\emptyset$. By Lemma 2, X and Y are d-separated given **T**. Every vertex in **S** is an ancestor of X or Y. Every vertex in **T** is either a parent of X, a vertex in **S**, or a parent of a vertex in **S**, hence $\mathbf{T} \subseteq \text{Ancestors}\{X,Y\}$. Moreover, every vertex in **T** is either a parent of X, a child of X, or a parent V of some vertex C in **S**, s.t. X→C. Any vertex in the first two categories is clearly adjacent to X. Any vertex in the last category is adjacent to X if C is an ancestor of X. Since C is in **S**, C is an ancestor of X or Y.

If X is not an ancestor of Y, then no child C of X is an ancestor of Y, so C is an ancestor of X; hence any parent V of C is also adjacent to X. ∴

**Lemma 3:** If A and B are not adjacent, then either A and B are d-separated by a set $\mathbf{T_A}$ of vertices adjacent to A or by a set $\mathbf{T_B}$ of vertices adjacent to B.

**Proof:** By Corollary B to Lemma 2, if A and B are not adjacent, then A and B are d-separated given $\mathbf{T_A}$ where:

$\mathbf{S_A} = \text{Children}(A) \cap \text{Ancestors}(\{A,B\})$

$\mathbf{T_A} = (\text{Parents}(\mathbf{S} \cup \{A\}) \cup \mathbf{S}) \setminus$
    $(\text{Descendants}(\text{Children}(A) \cap \text{Children}(B)) \cup \{A,B\})$.

**Case 1:** A is not an ancestor of B. From the Corollary B to Lemma 2, since A is not an ancestor of B, $\mathbf{T_A} \subseteq \{X \mid X \text{ adjacent to } A\}$.

**Case 2:** B is not an ancestor of A. It follows again by symmetry that A and B are d-separated given $\mathbf{T_B}$, where $\mathbf{T_B}$ is defined symmetrically to $\mathbf{T_A}$ in Case 1.

**Case 3:** B is an ancestor of A and A is an ancestor of B.

Now any vertex V in $\mathbf{T_A}$ is either a child of A, a parent of A or a parent of some vertex C in $\mathbf{S_A}$, s.t. A→C. Clearly vertices in the first two categories are adjacent to A; as before, vertices in the last category are adjacent to A if C is an ancestor of A. Any vertex in $\mathbf{S_A}$ is an ancestor of A or B. Since A is an ancestor of B, and B is an ancestor of A, it follows that every vertex in $\mathbf{S_A}$ is an ancestor of A, hence every vertex in $\mathbf{T_A}$ is adjacent to A. ∴

Let $G$ be any graph satisfying the Global Markov and Faithfulness conditions w.r.t. the distribution $P$ given as input. To find a set which d-separates some pair of variables A and B in $G$ the algorithm tests subsets of the vertices adjacent to A in $\Psi$, and subsets of vertices adjacent to B in $\Psi$ to see if they d-separate A and B. Since the vertices which are adjacent to A and B in $G$ are at all times a subset of the vertices adjacent to A and B in $\Psi$[16] Lemma 3 implies that step ¶A is guaranteed to find a set which d-separates A and B, if any set d-separates A and B in $G$.

### 4.2 SECTION ¶B

**Lemma 4:** Suppose that Y is not an ancestor of X or Z or a set **R**. If there is a set **S**, $\mathbf{R} \subset \mathbf{S}$, such that $Y \in \mathbf{S}$ and every proper subset **T** s.t. $\mathbf{R} \subsetneq \mathbf{T} \subset \mathbf{S}$, not containing Y, d-connects X and Z then **S** d-connects X and Z.

**Proof:** Let $\mathbf{T}^* = \text{Ancestors}(\{X,Z\} \cup \mathbf{R}) \cap \mathbf{S}$. Now, $\mathbf{R} \subset \mathbf{T}^*$, and $\mathbf{T}^*$ is a proper subset of **S**, so by hypothesis there is a d-connecting path, **P**, conditional on $\mathbf{T}^*$. By the definition of a d-connecting path, every element on **P** is either an ancestor of one of the endpoints, or $\mathbf{T}^*$. Moreover, by definition, every element in $\mathbf{T}^*$ is an ancestor of X or Z or **R**. Thus every element on the path **P** is an ancestor of X or Z or **R**. Since neither Y nor any element in $\mathbf{S} \setminus \mathbf{T}^*$ is an ancestor of X or Z or **R**, it follows that no vertex in $\mathbf{S} \setminus \mathbf{T}^*$ lies on **P**. Since $\mathbf{T}^* \subset \mathbf{S}$ the only way in which **P** could fail to d-connect given **S** would be if some element of $\mathbf{S} \setminus \mathbf{T}^*$ lay on the path (every collider active given $\mathbf{T}^*$ will remain active given **S**). Hence **P** still d-connects X and Z given **S**. ∴

**S** is said to be a *minimal d-separating* set for X and Y if X and Y are d-separated given **S**, and are d-connected given any proper subset of **S**.

**Corollary:** If **S** is a minimal d-separating set for X and Y, then any vertex in **S** is an ancestor of X or Y.

**Proof:** Follows by contraposition from Lemma 4 with $\mathbf{R} = \emptyset$. ∴

This shows that the unshielded non-collider orientation rule in ¶B is correct: If A and B, and B and C are adjacent, but **Sepset**(A,C) contains B, then by the nature of the search procedure A and C are not d-separated given any subset of **Sepset**(A,C) hence it follows that B is an ancestor of A or C, hence A∗−∗B∗−∗C should be oriented as A∗−∗<u>B</u>∗−∗C.

I will make frequent use of the following Lemma, which I state here without proof (It is a simple extension to the cyclic case of Lemma 3.3.1 in Spirtes *et al.*, 1993, p.376) The Lemma gives conditions under which a set of d-connecting paths may be joined to form a single d-connecting path.

---

[16] This is because if a pair of vertices X,Y are adjacent in $G$ then no set is found which d-separates them hence the edge between X and Y in $\Psi$ is never deleted.

**Lemma 3.3.1+**   (Richardson 1994, p.82)

In a directed (cyclic or acyclic) graph $G$ over a set of vertices $V$, *IF* $R$ is a sequence of distinct vertices in $V$ from A to B, $R \equiv <A \equiv X_0, \ldots X_{n+1} \equiv B>$, $S \subseteq V \setminus \{A,B\}$ and T is a set of undirected paths such that

(i) for each pair of consecutive vertices in $R$, $X_i$ and $X_{i+1}$, there is a unique undirected path in T that d-connects $X_i$ and $X_{i+1}$ given $S \setminus \{X_i, X_{i+1}\}$.

(ii) if some vertex $X_k$ in $R$, is in $S$, then the paths in T that contain $X_k$ as an endpoint collide at $X_k$.

(iii) if for three vertices $X_{k-1}$, $X_k$, $X_{k+1}$ occurring in $R$, the d-connecting paths in T between $X_{k-1}$ and $X_k$, and $X_k$ and $X_{k+1}$, collide at $X_k$ then $X_k$ has a descendant in $S$.

*THEN* there is a path $U$ in $G$ that d-connects $A \equiv X_0$ and $B \equiv X_{n+1}$ given $S$.

**Lemma 5:** If A and B are adjacent, B and C are adjacent, and B is an ancestor of A or C, then A and C are d-connected given any set $S \setminus \{A,C\}$, s.t. $B \notin S$.

**Proof:** Without loss of generality, let us suppose that B is an ancestor of C. It is sufficient to prove that A and C are d-connected conditional on S. There are two cases to consider:

**Case 1:** Some (proper) descendant of B is in S. It follows from Lemma 1 and the adjacency of A and B, that given any set S, conditional on $S \setminus \{A,B\}$, there is a d-connecting path from A to B, and likewise a d-connecting path from B to C, conditional on $S \setminus \{B,C\}$. Since some descendant of B is in $S \setminus \{A,C\}$, but $B \notin S \setminus \{A,C\}$, it follows by Lemma 3.3.1+ that A and C are d-connected, since it does not matter whether or not the path from A to B and from B to C collide at B.

**Case 2:** No descendant of B is in S. Again by Lemma 1 there is a path d-connecting from A to B. Since no descendant of B is in S, the directed path $B \to \ldots \to C$ is also d-connecting. Since $B \notin S$, Lemma 3.3.1+ implies A and C are d-connected by S. ∴

It follows by contraposition from Lemma 5 that if A and B are adjacent, B and C are adjacent, A and C are d-separated given **Sepset**<A,C>, and $B \notin$ **Sepset**<A,C>, then B is not an ancestor of A or C, hence ¶B correctly orients $A*-*B*-*C$ as $A \longrightarrow B \longleftarrow C$.

### 4.3 SECTION ¶C

**Lemma 6:** Suppose X is an ancestor of Y. If there is a set S such that A and Y are d-separated given S, X and Y are d-connected given S, and $X \notin S$, then A and X are d-separated given S, and some subset $T \subseteq S$ is a minimal d-separating set for A and X.

**Proof:** Let X be an ancestor of Y. Let S be any set s.t. there is a path $Q$ which d-connects X and Y given S, $X \notin S$, and A and Y are d-separated by S. Suppose, for a contradiction, that there is a path $P$ d-connecting A and X given S. There are now two cases:

**Case 1:** Some descendant of X is in S. Since $X \notin S$, and some descendant of X is in S, Lemma 3.3.1+ implies that the d-connecting paths $P$ and $Q$, can be joined to form a path d-connecting A to Y given $S$, a contradiction.

**Case 2:** No descendant of X is in S. In this case since X is an ancestor of Y, there is a d-connecting directed path $Q^*$, $X \to \ldots \to Y$, given S. By Lemma 3.3.1+ $P$ and $Q^*$ can be joined to form a path d-connecting A and Y given S, a contradiction.

Thus under the conditions in the antecedent, S is a d-separating set for A and X. Let $T$ be the smallest subset of $S$ which d-separates A and X, $T$ is a minimal d-separating set for A and X. ∴

**Lemma 7:** Let A, X and Y be three vertices in a graph, s.t. X and Y are adjacent. If there is a set S s.t. $X \notin S$, A and Y are d-separated given S, while A and X are d-connected given S, then X is not an ancestor of Y.

**Proof:** If X and Y are adjacent then X and Y are d-connected by every set S, s.t. $X, Y \notin S$. If there is a set S which d-separates A and Y, but does not contain any subset which d-separates A and X, where X is adjacent to Y, and $X \notin S$, then S does not contain a (minimal) d-separating set for A and X, hence, by Lemma 6, X is not an ancestor of Y. ∴

¶C simply applies Lemma 7: If A and X are d-connected given **Sepset**<A,Y>, and $X \notin$ **Sepset**<A,Y>, then since **Sepset**<A,Y> d-separates A and Y, by Lemma 7, ¶C correctly orients $X*-*Y$ as $X \longleftarrow Y$.

### 4.4 SECTION ¶D

**Lemma 8:** If in a graph $G$, Y is a descendant of a common child of X and Z, then X and Z are d-connected by any set S s.t. $Y \in S$, $X, Z \notin S$.

**Proof:** If Y is a descendant of a common child C of X and Z, then the path $X \to C \leftarrow Z$ d-connects X and Z given any set S, s.t. $Y \in S$, $X, Z \notin S$. ∴

**Corollary:** If in a graph $G$, X and Y are adjacent, Y and Z are adjacent, but X and Z are not adjacent, Y is not an ancestor of X or Z, and there is some set S such that $Y \in S$, and X and Z are d-separated given S, then Y is not a descendant of a common child of X and Z.

**Lemma 9:** If in graph $G$, Y is not a descendant of a common child of X and Z, then X and Z are d-separated by the set $T$, defined as follows:

$S = $ Children(X) $\cap$ Ancestors({X,Y,Z})

$T = ($Parents$(S \cup \{X\}) \cup S) \setminus$
(Descendants(Children(X)$\cap$Children(Z)) $\cup \{X,Z\}$)

Further, if X and Y, and Y and Z are adjacent then $Y \in T$.

**Proof:** Lemma 2, with $Q=\{Y\}$ implies that X and Z are d-separated by T. If Y is a child of X, then since Y is an ancestor of Y, $Y \in S$. Since Y is not a descendant of a common child of X and Z, $Y \in T$. If Y is a parent of X, then since Y is not a descendant of a common child of X and Z, $Y \in T$. If X and Y have a common child C that is an ancestor of X or Y, then $C \in S$; since Y is a parent of C, and Y is not a descendant of a common child of X and Z then $Y \in T$. So if X and Y are adjacent then $Y \in T$. ∴

¶D considers each triple A—>B<—C in Ψ, A and C are not adjacent, in turn, and tries to find a set R ⊆ Local(Ψ,A)\{B,C} s.t. A and C are d-separated by R ∪ {B}. If A and C are d-separated by a set containing B, then Lemma 8 implies that B is not a descendant of a common child of A and C. It then follows from Lemma 9 that the set T in Lemma 9 is s.t. B∈T, A and C are d-separated by T, and T ⊆ Local(Ψ,X). So ¶D will find a set which d-separates A and C, but contains B, if such a set exists.

### 4.5 SECTION ¶E

**Lemma 10:** If in a graph G, A and D are adjacent, D and C are adjacent, A and C are not adjacent, D is an ancestor of B, then any set S such that B∈S, and A and C are d-separated by S, also contains D.

**Proof** Suppose for a contradiction that A and C were d-separated by a set S, s.t. B∈S, D∉S. Since A is adjacent to D, (D,A∉S), by Lemma 1 there is an undirected path P d-connecting A and D given S. Likewise there is a path Q d-connecting D and C given S. Since D is an ancestor of B, B∈S, but D∉S, Lemma 3.3.1+ implies that P and Q can be joined to form a new path d-connecting A and C given S. This is a contradiction.∴

By contraposition Lemma 10 justifies ¶E in the case where A—>B<—C, A—>D<—C, D∉Supset<A,B,C>, and so B*—*D is oriented as B—>D.

In the case in which A—>B<—C, A—>D<—C, and D∈Supset<A,B,C> Lemma 4, and the nature of the search for Supset<A,B,C>[17] imply that D is an ancestor of {A,B,C}. But since there are arrowheads at D on the edges A—>D<—C, D is not an ancestor of A or C, so D is an ancestor of B. Hence, ¶E correctly orients B*—*D as B*—D.

In the case where A—>B<—C, A→D←C in Ψ, (A and C not adjacent and no dotted line A—>D<—C), Lemma 8 implies that, since A and C are d-connected by any set S s.t. D∈S, (A,C∉S), D is a descendant of a common child of A and C. Since A and C are d-separated by Supset<A,B,C>, and B∈Supset<A,B,C>, then B is not a descendant of D. So ¶E correctly orients B*—*D as B<—D.

### 4.6 SECTION ¶F

**Lemma 11:** If X and Z are d-separated by some set R, then for all sets Q ⊆ Ancestors(R ∪ {X,Z})\{X,Z}, X and Z are d-separated by R ∪ Q.

**Proof:** Suppose for a contradiction that there is a path P d-connecting X and Z given R ∪ Q. It follows that every vertex on P is an ancestor of either X, Z, or R ∪ Q. Since Q ⊆ Ancestors(R ∪ {X,Z}) it follows that every vertex on P is an ancestor of X, Z or R.

Let A be the collider furthest from X on P which is an ancestor of X and not R (or X if no such collider exists),

---

[17] Namely, that ¶D looks for the smallest set containing B, which d-separates A and C.

---

let B be the first collider after A on P which is an ancestor of Z and not R (or Z if no such exists). The paths X←...←A, and B→...→Z are d-connecting given R, since no vertex on the paths is in R. The subpath of P between A and B is also d-connecting given R since every collider is an ancestor of R, and no non-collider lies in R, since, by hypothesis P d-connects given R ∪ Q. Lemma 3.3.1+ implies that these three paths can be joined to form a path d-connecting X and Z given R. This is a contradiction.∴

In ¶F, since A and C are d-separated by Supset<A,B,C> ⊇ {B}, by Lemma 11, if A and C are d-connected given Supset<A,B,C> ∪ {D}, then D is not an ancestor of B. Further, since B and D are adjacent, B is an ancestor of D. So ¶F correctly orients B*—*D as B—>D in Ψ.

This completes the proof of the correctness.    ∴

### Acknowledgements

I thank P. Spirtes, C. Glymour, R. Scheines & C. Meek for helpful conversations. Research supported by NSF grant 9102169.

### References

BOLLEN, K. (1989) Structural Equations with Latent Variables, Wiley, NY.

KOSTER, J.T.A.(1994) Markov Properties of Non-Recursive Causal Models. MS.

LAURITZEN, S, DAWID, A., LARSEN, B., LEIMER, H., (1990). Independence properties of directed Markov fields, *Networks* **20**

PEARL, J. (1988). Probabilistic Reasoning in Intelligent Systems. Morgan Kaufmann, San Mateo, CA.

PEARL, J. (1993). On the Statistical Interpretation of Structural Equations. Technical Report R-200, Cognitive Systems Lab., UCLA.

RICHARDSON, T. (1994). Properties of Cyclic Graphical Models. MS Thesis, Carnegie Mellon University

RICHARDSON, T. (1996). Discovering Cyclic Causal Structure. Tech Report CMU-PHIL 68.

SPIRTES, P. (1995). Directed Cyclic Graphical Representations of Feedback Models. *Proceedings UAI 95*, ed. P. Besnard and S. Hanks, Morgan Kaufmann CA.

SPIRTES, P., GLYMOUR , C. & SCHEINES , R. (1993) Causation, Prediction & Search. Lecture Notes in Statistics, Springer-Verlag.

VERMA, T. & PEARL, J., (1990). On Equivalence of Causal Models. Technical Report R-150, Cognitive Systems Lab., UCLA.